\documentclass[sigconf]{acmart}

\usepackage{booktabs} 
\usepackage{tabularx}

\copyrightyear{2019}
\acmYear{2019}
\setcopyright{acmlicensed}
\acmConference[GECCO '19]{Genetic and Evolutionary Computation Conference}{July 13--17, 2019}{Prague, Czech Republic}
\acmBooktitle{Genetic and Evolutionary Computation Conference (GECCO '19), July 13--17, 2019, Prague, Czech Republic}
\acmPrice{15.00}
\acmDOI{10.1145/3321707.3321776}
\acmISBN{978-1-4503-6111-8/19/07}

\begin{document}

\title{Semantic variation operators for multidimensional genetic programming}

\author{William La Cava}
\authornote{Corresponding Author}
\orcid{1234-5678-9012}
\affiliation{%
  \institution{University of Pennsylvania}
  \streetaddress{3700 Hamilton Walk}
  \city{Philadelphia} 
  \state{PA} 
  \postcode{19104}
}
\email{lacava@upenn.edu}

\author{Jason H. Moore}
\affiliation{%
  \institution{University of Pennsylvania}
  \streetaddress{3700 Hamilton Walk}
  \city{Philadelphia} 
  \state{PA} 
  \postcode{19104}
}
\email{jhmoore@upenn.edu}


\begin{abstract}
     Multidimensional genetic programming represents candidate solutions as sets of programs, and thereby provides an interesting framework for exploiting building block identification. 
     Towards this goal, we investigate the use of machine learning as a way to bias which components of programs are promoted, and propose two semantic operators to choose where useful building blocks are placed during crossover.  
     A forward stagewise crossover operator we propose leads to significant improvements on a set of regression problems, and produces state-of-the-art results in a large benchmark study.
     We discuss this architecture and others in terms of their propensity for allowing heuristic search to utilize information during the evolutionary process. 
     Finally, we look at the collinearity and complexity of the data representations that result from these architectures, with a view towards disentangling factors of variation in application. 
\end{abstract}

%
%

\keywords{representation learning, feature construction, variation, regression}

\maketitle

\section{Introduction}\label{s:intro}
A central theme in genetic programming (GP) is how to identify, propagate, and properly compose the components of programs that contribute to good solutions.
In the context of classification and regression, these building blocks fill the role of ``feature engineering".
That is to say, building blocks of GP solutions are meant to explain the underlying factors of variation that produce the observed response. 
The task of optimizing a set of explanatory features for a problem is known as {\it representation learning}, especially in the larger machine learning (ML) community~\cite{bengio_representation_2013}. 
Representation learning is a fundamental challenge in ML due to its computational complexity and the role the representation plays in model accuracy and interpretation. 
Interestingly, a variant of GP known as multidimensional GP (MGP) makes this relationship between building block discovery and representation learning explicit by optimizing a set of programs, each of which is an independent feature in the ML model. 
Our goal in this paper is to introduce semantic variation methods to MGP, with the goal of improving the representations it produces.  

What makes a representation good? 
At the minimum, a good representation produces a model with better generalization than a model trained only on the raw data attributes. 
In addition, a good representation teases apart the factors of variation in the data into independent components. 
Finally, an ideal representation is succinct so as to promote intelligibility. In other words, a representation should only have as many features as there are independent factors in the process. 
Our discussion centers around these three motivations.

In the following section, we attempt to summarize the large body of work concerning feature construction / representation learning in GP, especially those methods that use ML to promote building blocks. 
This provides context for the MGP family of methods. 
We then describe our main contribution: the proposed methods of crossover in Section~\ref{s:methods}. 
We conduct an experiment at first on 8 regression problems, considering full hyperparameter tuning, and analyze the representations that are produced with and without the new crossover methods. 
Finally, we benchmark the new methods against many ML and GP methods on more than 100 open source regression problems. 
We find that the new methods of crossover lead to state-of-the-art results for regression.
Our discussion points to further directions for improving representation quality within this framework. 

\section{Background}\label{s:back}

Feature construction / representation learning has been a consistent theme in the GP community and has been studied with various architectures. 
Without major changes, GP can be applied to the task of identifying single features for regression and classification (or multiple features in the multiclass case~\cite{neshatian_filter_2012}), and this approach has been explored in several works, summarized in~\cite{muharram_evolutionary_2005}.
These filter approaches generally use an information-theoretic measure to determine how good a program is likely to be as a feature. 
Optimizing single features requires basically no changes to GP's methodology, but lacks the power to optimize features for the multivariate context in which they are typically used. 

Another approach that has been studied is to treat each individual in the population as a feature, and to optimize an ensemble model of the entire population~\cite{de_melo_kaizen_2014, arnaldo_building_2015,veloso_de_melo_automatic_2017, la_cava_general_2017,la_cava_ensemble_2017}. 
Only a single regression model is trained per generation, which demands minimal overhead. 
However, it is not well known how to properly select and vary features evolved by such a process. 
Since each individual is a feature, its fitness is heavily dependent on the current population. 
Furthermore, a desirable set of features should be essentially orthogonal in order to create a well-conditioned representation, and convergent evolutionary processes aim in the opposite direction.
To overcome issues of collinearity and a convergent search process, the following ideas have been proposed. 
In evolutionary feature synthesis (EFS)~\cite{arnaldo_building_2015}, features are selected proportionally to their coefficient in a regularized linear model; in order to prevent multicollinearity, correlation thresholds are implemented during variation to keep children different from their parents. 
In the feature engineering wrapper (FEW)~\cite{la_cava_ensemble_2017,la_cava_general_2017}, multicollinearity is selected against by using a survival version of $\epsilon$-lexicase selection to choose features. 
In Kaizen GP~\cite{de_melo_kaizen_2014, veloso_de_melo_automatic_2017}, individuals are only added to the model if they pass a significance test, in a hill climbing fashion.
Another option is to not use an evolutionary updating scheme at all, but rather to create a large set of random features and fit an ML model to this, as in FFX~\cite{mcconaghy_ffx:_2011}. 
More recently, Vanneschi et. al. explored one step linear combinations of random programs~\cite{vanneschi_psxo:_2017}, experimentally showing that they often lead to overfitting. 

Rather than building a model from the entire population, one could apply an ML method to the entire program trace as a means of identifying building blocks~\cite{krawiec_behavioral_2016}. 
Multiple regression GP (MRGP)~\cite{arnaldo_multiple_2014} defines a program's behavior as the Lasso~\cite{tibshirani_regression_1996} estimate generated over the entire program's trace.
One downside of this approach is the likely presence of highly correlated features in the program trace, leading to an ill-conditioned regression matrix. 
In a similar vein to MRGP, Behavioral GP~\cite{krawiec_behavioral_2014} extracts information from the entire program trace, this time using a decision tree algorithm to identify important building blocks, which are stored in an archive for re-use. 
In both algorithms, the key insight is to use ML with program traces to undo the complex masking effect that program execution has on the behavior of building blocks that are downstream from other operations in the program (for further discussion on the topic of program traces see~\cite{krawiec_relationships_2012}). 

MGP is a framework that is, in some sense, in between the ensemble techniques and the program trace techniques described above.
Programs are represented as sets of separate subprograms, usually trees.
Unlike population-wide models, the fitness of each individual is directly related to its model predictions, and individuals in the population benefit from typical evolutionary optimization processes. 
Unlike program trace-based methods, by using multi-output programs, MGP exposes {\it independent components} of the total program behavior to the ML process that produces the model.  
As a result, building blocks are easier to isolate and share among the population in direct ways.

Examples of MGP include Krawiec's method~\cite{krawiec_genetic_2002}, M2GP~\cite{ingalalli_multi-dimensional_2014}, M3GP~\cite{munoz_m3gpmulticlass_2015}, e-M3GP~\cite{silva_multiclass_2015}, M4GP~\cite{la_cava_multidimensional_2018}, and FEAT~\cite{la_cava_learning_2019}.
In all of these methods, individuals in the population produce a set of corresponding outputs that are then fed into a deterministic ML method to produce the program's regression or classification estimates. 
In the case of M2GP, M3GP, and M4GP, classification proceeds using a nearest centroid classifier~\cite{tibshirani_diagnosis_2002}, whereas linear regression methods are used for regression with M3GP~\cite{munoz_evolving_2018} and FEAT. 
Although a number of methods have been proposed in the MGP paradigm, they have not made much use of the semantics of independent building blocks in each program that this architecture creates. 
An exception is~\cite{la_cava_learning_2019}, in which the authors use the coefficient magnitude to weight probabilities of mutation. 
The main contribution of this study is the development of semantic crossover schemes to leverage the architecture of MGP to a larger degree than these previous studies.

\section{Methods}\label{s:methods}
 
In this paper we focus on the task of regression, with the goal of building a predictive model $\hat{y}(\mathbf{x})$ using $N$ paired examples $\mathcal{T} = \{(\mathbf{x}_i,y_i)\}_{i = 1}^{N}$. 
The regression model $\hat{y}(\mathbf{x})$ associates the inputs $\mathbf{x} \in \mathbb{R}^d$ with a real-valued output $y \in \mathbb{R}$. 
The goal of feature engineering / representation learning is to find a new representation of $\mathbf{x}$ via a $m$-dimensional feature mapping $\boldsymbol{\phi}(\mathbf{x}): \mathbb{R}^d \rightarrow \mathbb{R}^m$, such that the model $\hat{y}(\boldsymbol{\phi}(\mathbf{x}))$ outperforms the model $\hat{y}(\mathbf{x})$ by some pre-defined metric (see above).

In MGP, each individual in the population is a candidate representation, $\boldsymbol{\phi}(\mathbf{x})$, consisting of a set of programs $[ \phi_1,\;\dots\;,\phi_m]$. As an example, the individual  
\[
[ (+ \; (x_1) \; (x_2) ), \; ( \cos \;(x_3) \; ), \; (\exp \; (\text{cube} \; (x_1)))]
 \] 
would encode a representation with three features: $(x_1+x_2)$, $\cos (x_3)$, and $\exp (x_1^3)$.  
Throughout the paper, we refer to these subprograms $\phi$ as {\it features}, and use the word {\it attribute} to refer to the independent variables in $\mathbf{x}$.

MGP methods share this representation in common, and differ in terms of 1) the ML method used to generate the model prediction, i.e. $\hat{y}(\boldsymbol{\phi})$, 2) the crossover and mutation operators used, and 3) the selection process used. The crossover operators proposed in this section will work with any MGP method, but are designed with a linear ML pairing in mind.

\subsection{Feature Engineering Automation Tool}

We study a recent MGP method named the Feature Engineering Automation Tool (FEAT)~\cite{la_cava_learning_2019}, in which candidate features are parameterized by weights, $\boldsymbol{\theta}$, and used to fit a linear model 

\begin{equation}
    \hat{y} = \sum_{i=1}^m{\beta_i\phi_i(\mathbf{x}, \boldsymbol{\theta})} 
    \label{eq:ml}
\end{equation}

The coefficients $[\beta_1,\;\dots,\;\beta_m]$ are determined using ridge regression~\cite{hoerl_ridge_1970}.
Note that each $\phi$ is normalized to zero mean, unit variance before ridge regression is applied.
The parameters $\boldsymbol{\theta}$ are attached to the edges of differentiable operators and updated each generation via gradient descent. 
The fitness of each individual in FEAT is its mean squared error (MSE) on the training set.

\paragraph{Feedback}
In order to promote building blocks, FEAT uses feedback from the ML process to bias the variation step. 
In a nutshell, the probability of a feature in $\boldsymbol{\phi}$ being mutated or replaced by crossover is inversely related the magnitude of its coefficient $\beta$ in Eqn.~\ref{eq:ml}.  
Let $\tilde{\beta}_i(n) = |\beta_i|/\sum_i^m|\beta_i|$. 
The normalized coefficient magnitudes $\tilde{\beta} \in [0,1]$ are used to define softmax-normalized probabilities.
The probability of mutation for feature $i$ in program $n$ is denoted $PM_i(n)$, and defined as follows: 

\begin{eqnarray}\label{eq:mutate}
    s_i(n) &=&  \exp(1-\tilde{\beta}_{i}) / \sum_i^m{\exp(1-\tilde{\beta}_i)} \nonumber \\
    PM_i(n) &=&  \gamma s_i(n) + (1-\gamma) \frac{1}{m} 
\end{eqnarray}
 
Here, $\gamma$ is a parameter that controls the amount of feedback from the weights that is used to bias the selection of feature $i$ for mutation. In our experiments, we tune $\gamma$, and also test whether the softmax normalization of $s_i(n)$ is useful.

\paragraph{Selection}
In the initial work introducing FEAT~\cite{la_cava_learning_2019}, the authors compared several optimization schemes, including NSGA-II, simulated annealing, random search, $\epsilon$-lexicase selection~\cite{la_cava_epsilon-lexicase_2016}, and a hybrid of $\epsilon$-lexicase selection with NSGA-II's survival scheme. 
The final combination achieved the best results in the analysis, and is used in our work here. 
The original work attempted to address the issue of disentanglement by measuring the multicollinearity of representations and setting this metric as an additional objective during survival. 
However, none of the objectives tried resulted in less correlated final representations, and actually made them slightly worse.  
One of the goals of this paper is to explore a second hypothesis, that disentangled representations can be encouraged through the use of semantic variation operators, rather than through additional objectives.

\paragraph{Variation}
Semantic variation operators have not yet been proposed for MGP frameworks. 
The most recent MGP techniques (M3GP, M4GP and FEAT) adopt special variation operators that vary programs at the feature level.  
Feature crossover (called ``root crossover" in~\cite{munoz_m3gpmulticlass_2015}) swaps features between two parent representations. 
Feature mutation may delete a feature or add a random feature. 
M3GP, M4GP and FEAT also use standard subtree crossover and point mutation operators, and each of these operators occur with equal probability. 
In FEAT, however, the probabilities of each feature being chosen for mutation or crossover is determined by Eqn.~\ref{eq:mutate}. 

In the following section, we describe two new semantic crossover operators for MGP that attempt to maintain orthogonality in the representations while moving towards models with lower residuals. 

\subsection{Semantic Crossover}
The following two crossover methods are called {\it semantic} because they use information about the program's outputs to determine the recombination that occurs to produce a child from two parents. 
Both operators are based on the following observations. 
We have two parent representations, $\boldsymbol{\phi}_{p1}$ and $\boldsymbol{\phi}_{p2}$, with corresponding model outputs $\hat{y}_{p1}$ and $\hat{y}_{p2}$ that are linear combinations of their respective representations, as in Eqn.~\ref{eq:ml}.
We want to produce the best combination of $\boldsymbol{\phi}_{p1}$ and $\boldsymbol{\phi}_{p2}$ for the child representation $\boldsymbol{\phi}_{c}$. 
Basically we can treat this as a feature selection problem, where we have features $\boldsymbol{\phi}_A = \boldsymbol{\phi}_{p1} \cup \boldsymbol{\phi}_{p2}$ and we want to pick the best. 
On one hand we could simply concatenate the feature sets, and generate a new model $\hat{y}(\boldsymbol{\phi}_A)$, which is the linear model fit to all features of both parents. 
This approach would lead to exponential growth in offspring, which would run against our goal of lowering complexity. 

In lieu of that approach, we propose here what are essentially regularized versions of semantic crossover that constrain the number of features in the offspring to be of equal cardinality to $\boldsymbol{\phi}_{p1}$, i.e. $|\boldsymbol{\phi}_c| = |\boldsymbol{\phi}_{p1}|$. 
The first operator, best residual fit crossover (ResXO), chooses a feature from $\boldsymbol{\phi}_{p1}$ to be replaced, and then chooses the feature in $\boldsymbol{\phi}_{p2}$ that best approximates the residual of the model after removing this feature. 
The second operator, stagewise crossover (StageXO), uses forward stagewise regression~\cite{james_introduction_2013} as a feature selection method to iteratively construct the offspring.  

\subsection{Best residual fit crossover (ResXO)}
Given parents $p1$ and $p2$, ResXO swaps a feature in $p1$ with the feature in $p2$ that most closely approximates the residual error of $p1$ with the selected feature removed. The child representation is denoted as $\boldsymbol{\phi}_c$. The steps are as follows:

\begin{enumerate}
    \item pick $\phi_d$ from $\boldsymbol{\phi}_{p1}$ using probabilities given by Eqn.~\ref{eq:mutate}.  
    \item calculate the residual of $p1$ without $\phi_d$: \[r = y - \hat{y}_{p1} - \beta_d\phi_d\]
    \item choose $\phi^*$ from $\boldsymbol{\phi}_{p2}$, which is the feature most correlated with $r$.
    \item $\boldsymbol{\phi}_c = \boldsymbol{\phi}_{p1}$ with $\phi_d$ replaced by $\phi^*$.
\end{enumerate}

ResXO is a semantic backpropagation operator~\cite{ffrancon_memetic_2015, pawlak_semantic_2015,graff_semantic_2015}, since it seeks to replace a component of the parent program with a subprogram most closely matching the desired semantics, given by $r$. 
Within the MGP framework, this backpropagation is very simple, and does not require complex inversion operations to be introduced. 
We expect that ResXO will also lead to lower correlations between features in $\boldsymbol{\phi}_c$ than in $\boldsymbol{\phi}_{p1}$. 
To understand why, consider that \[r = y - \sum_{\phi_i \in \boldsymbol{\phi}_{p1} \setminus \phi_d} \beta_i \phi_i\] 
Therefore $r$ should have low correlation with the rest of the $p1$'s representation. 
Assuming the replacement feature from $\boldsymbol{\phi}_{p2}$ closely matches $r$, it should also be uncorrelated with $\{\boldsymbol{\phi}_{p1} \setminus \phi_d\}$.
Note that ResXO may produce an individual with higher squared error than its parents, since $\phi_d$ may be more correlated with $r$ than $\phi^*$. 

\subsection{Forward stagewise crossover (StageXO)}
Rather than restricting crossover to the replacement of a single feature, the crossover operator can be used to compile the set of features that iteratively reduce the target error using a forward stagewise crossover method we call StageXO. The procedure is as follows:

\begin{enumerate}
    \item set the initial residual equal to the target: $r = y$. Center means around zero for all $\phi$.
    \item set $\boldsymbol{\phi}_A$ to be all subprograms in $\boldsymbol{\phi}_{p1}$ and $\boldsymbol{\phi}_{p2}$.
    \item while $|\boldsymbol{\phi}_{c}| < |\boldsymbol{\phi}_{p1}|$:
    \begin{enumerate}
        \item pick $\phi^*$ from $\boldsymbol{\phi}_{A}$ which is most correlated with $r$.
        \item compute the least squares coefficient $b$ for $\phi^*$ fit to ${r}$.
        \item update ${r} = r - b\phi^*$
        \item add $\phi^*$ to $\boldsymbol{\phi}_c$.
        \item remove $\phi^*$ from $\boldsymbol{\phi}_A$.
    \end{enumerate}
\end{enumerate}

Unlike feature selection methods like forward/backward stepwise selection, forward stagewise selection only calculates the weight of a single feature at a time, and is thus more lightweight. 
The downside of this approach in the context of regression is that it generally takes more iterations to reach the least squares coefficients of the complete model~\cite{friedman_elements_2001}. 
In our case this is unimportant, since we are only interested in quickly choosing the most important features, which are then used to fit a multiple linear regression model. 
We expect the child representation returned by StageXO to contain uncorrelated features since the residual is updated each iteration to remove the portion of the response explained by previous features.

Forward stagewise regression, and therefore the StageXO operator, is closely related to boosting~\cite{freund_desicion-theoretic_1995}. 
In both cases the residual is iteratively reduced by adding model components (weak learners in the case of boosting, and features/building blocks in our case).
The relationship between forward stagewise regression, boosting, and regularized linear models is expounded upon in~\cite{friedman_elements_2001}. 
The stagewise additive modeling paradigm is also used by a recent GP technique called Wave~\cite{medernach_new_2016}, in which GP runs are iteratively trained on residuals of previous runs. 
The insight here is that the unique representation of programs in MGP allows the same general methodology to be exploited for combining partial solutions during crossover, rather than as a post-run ensemble method.  
\section{Experiment}\label{s:exp}
Our experiment consists of two stages. First, we conduct an extensive study of FEAT with and without the semantic crossover operators introduced in Section~\ref{s:methods}. In this study we vary the hyperparameters related to variation and analyze the results in detail for 8 regression problems. In the second study, we apply FEAT, FEATResXO, and FEATStageXO to more than 100 problems from the PMLB regression benchmark~\cite{orzechowski_where_2018}. These variants are compared to state-of-the-art symbolic regression and ML methods.  

\subsection{Hyperparameter study}
Despite several MGP methods having been proposed, there has not been a systematic study of the effect of variation operators on the performance of this family of methods. 
To fill this gap, and to properly analyze the new methods introduced in this paper, we performed a grid search of variation hyperparameters on 8 regression problems. 
The hyperparameters that were varied are shown in Table~\ref{tbl:settings}.

\begin{table}
    \centering
    \caption{Hyperparameter values for FEAT in the experiments. The bottom values are fixed.}\label{tbl:settings}
    \begin{tabularx}{\columnwidth}{X r} \hline
        Hyperparameter & Values \\ \hline
        probability of crossover (complement: mutation)  &   [0,0.25,0.5,{ 0.75},1.0]  \\
        feedback ($\gamma$, Eqn.~\ref{eq:mutate})    &   [0, { 0.25}, 0.5, 0.75, 1.0]   \\
        type of feature crossover   &   [Standard, ResXO, StageXO]  \\
        probability of feature crossover (complement: subtree crossover) &   [0.5, 0.75, 1.0]    \\
        feedback softmax normalization  &   [On, { Off}]   \\
        population size &   500 \\
        generations     &   100 (200 for PMLB) \\
        max depth       &   6   \\
        maximum dimensionality  & min(50, $2*|\mathbf{x}|$)  \\
        iterations of gradient descent & 10 \\ \hline
    \end{tabularx}

\end{table}

Feedback softmax normalization refers to the softmax transformation in Eqn.~\ref{eq:mutate}; we tested for whether this normalization, which assumes a multinomial distribution of probabilities, was useful. 
The eight comparison problems are listed in Table~\ref{tbl:regression}.

\begin{table}
\centering
\caption{Regression problems used for method comparisons.}\label{tbl:regression}
\begin{tabularx}{\columnwidth}{X r r } \toprule
Problem & Dimension & Samples \\ \midrule
Airfoil & 5	& 1503 \\
Concrete	& 	8	& 1030	\\
ENC & 8 & 768 \\
ENH & 8 & 768 \\
Housing & 14 & 506 \\
Tower & 25 & 3135 \\
UBall5D & 5 & 6024 \\ 
Yacht	& 6	&	309	\\ \midrule
\end{tabularx}
\end{table}

\subsection{Benchmark comparison}
In the second study, we compared FEAT with each crossover variant to 15 other methods: 5 GP methods~\cite{castelli_c++_2015, arnaldo_multiple_2014, schmidt_age-fitness_2011,la_cava_probabilistic_2018} and 10 ML methods from scikit-learn~\cite{pedregosa_scikit-learn:_2011}.  
The 5 GP methods we compared to are:
\begin{itemize}
    \item Geometric Semantic GP (GSGP)~\cite{castelli_c++_2015}
    \item MRGP~\cite{arnaldo_multiple_2014}
    \item Age-fitness Pareto Optimization (AFP)~\cite{schmidt_age-fitness_2011}
    \item $\epsilon$-lexicase selection (EPLEX)~\cite{la_cava_probabilistic_2018}
    \item $\epsilon$-lexicase selection with 1 million evaluations (EPLEX-1M)~\cite{la_cava_probabilistic_2018}
\end{itemize}

These methods were benchmarked on 94 open-source datasets collected in the Penn ML Benchmark~\cite{olson_pmlb:_2017}. 
We used results from Orzechowski et. al.'s benchmark analysis~\cite{orzechowski_where_2018} as a comparison, and followed the same validation procedure. 
Each comparison method underwent hyperparameter tuning using 5-fold cross validation on a 75\% split of the training set, and was then tested on a 25\% test fold. 
The hyperparameters are detailed in Table 1 of the original work~\cite{orzechowski_where_2018}. 
This process was repeated for 10 trials. 
GP methods were given 100,000 evaluations, apart from EPLEX-1M which used 1 million. 
For FEAT, we did not re-tune the hyperparameters, instead using the values determined from the hyperparameter tuning experiment.

\begin{figure}
    \centering
    \includegraphics[width=\columnwidth]{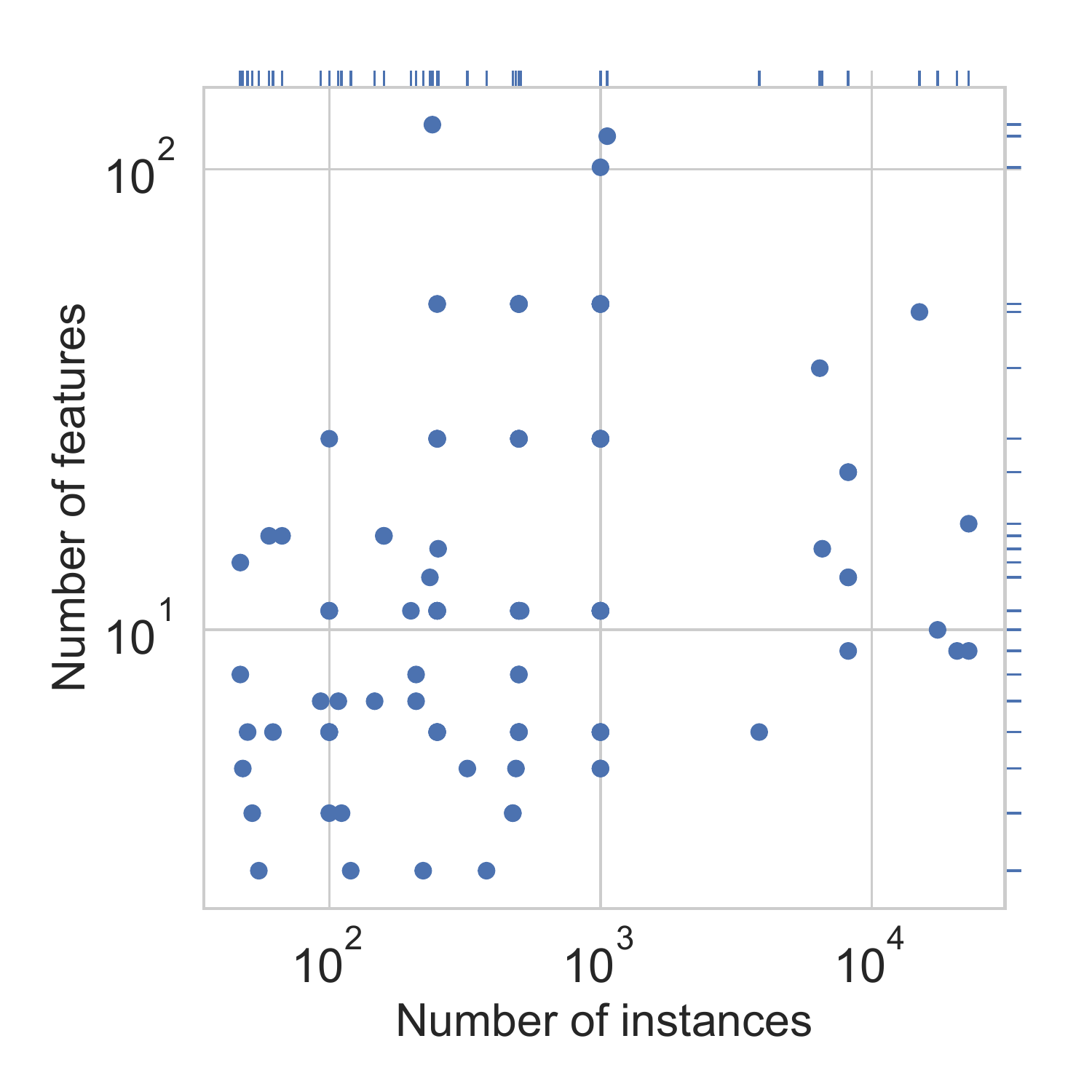}
    \caption{Properties of the benchmarks used from PMLB~\cite{olson_pmlb:_2017}.}\label{fig:pmlb}
\end{figure}

We also extended the PMLB comparison to larger datasets because we were interested in 1) how FEAT handled larger datasets (the original study was restricted to smaller datasets, c.f. Fig. 1 of \cite{orzechowski_where_2018}) and also how the size of the final models compared to other top-ranking algorithms such as XGBoost~\cite{chen_xgboost:_2016} and multilayer perceptron (MLP). 
The extended analysis included 111 datasets, whose properties are shown in Fig.~\ref{fig:pmlb}. 
These datasets are used to evaluate the complexity of the final models. 

\subsection{Metrics}
As mentioned earlier, we consider there to be three over-arching goals when learning a representation. 
The first is that $\boldsymbol{\phi}(x)$ leads to a model with a low generalization error.  
To measure this, we compare the mean squared error (MSE) and coefficient of determination ($R^2$) of each model output on the test set.
We also wish to minimize the complexity of the representation. 
To measure the complexity of solutions in FEAT, we count the total number of nodes in the final representation. 
For comparison to XGBoost, we count the number of nodes in the trees, and for comparison to MLP, we count the number of nodes in the network.
Finally, we want a representation that is ``disentangled", meaning that each feature of $\boldsymbol{\phi}$ is as orthogonal to the others as possible. 
One such measure is the pairwise Pearson correlations of features in $\boldsymbol{\phi}$, written as 

\begin{equation}
    Corr(\boldsymbol{\phi}) = \frac{1}{N(N-1)} \sum_{\phi_i,\phi_j \in \boldsymbol{\phi}, i \neq j}{\left( \frac{\text{cov}(\phi_i,\phi_j)}{\sigma(\phi_i)\sigma{(\phi_j)}} \right)^2} \label{eq:corr}
\end{equation}
 
We use this equation to compare the final representations across selection methods. 

\section{Results}
The hyperparameter tuning results are presented first. In addition to test score reporting, we plot various views of the data with respect to different hyperparameters, and also look at representation correlations in the resultant models and statistical comparisons. In the subsequent section, the PMLB comparison results are shown, including score comparisons, runtime comparisons, statistical tests and comparisons of the final model sizes for FEAT, XGBoost, and MLP learners. 
\subsection{Hyperparameter tuning}

\begin{figure}
    \centering
    \includegraphics[width=\columnwidth]{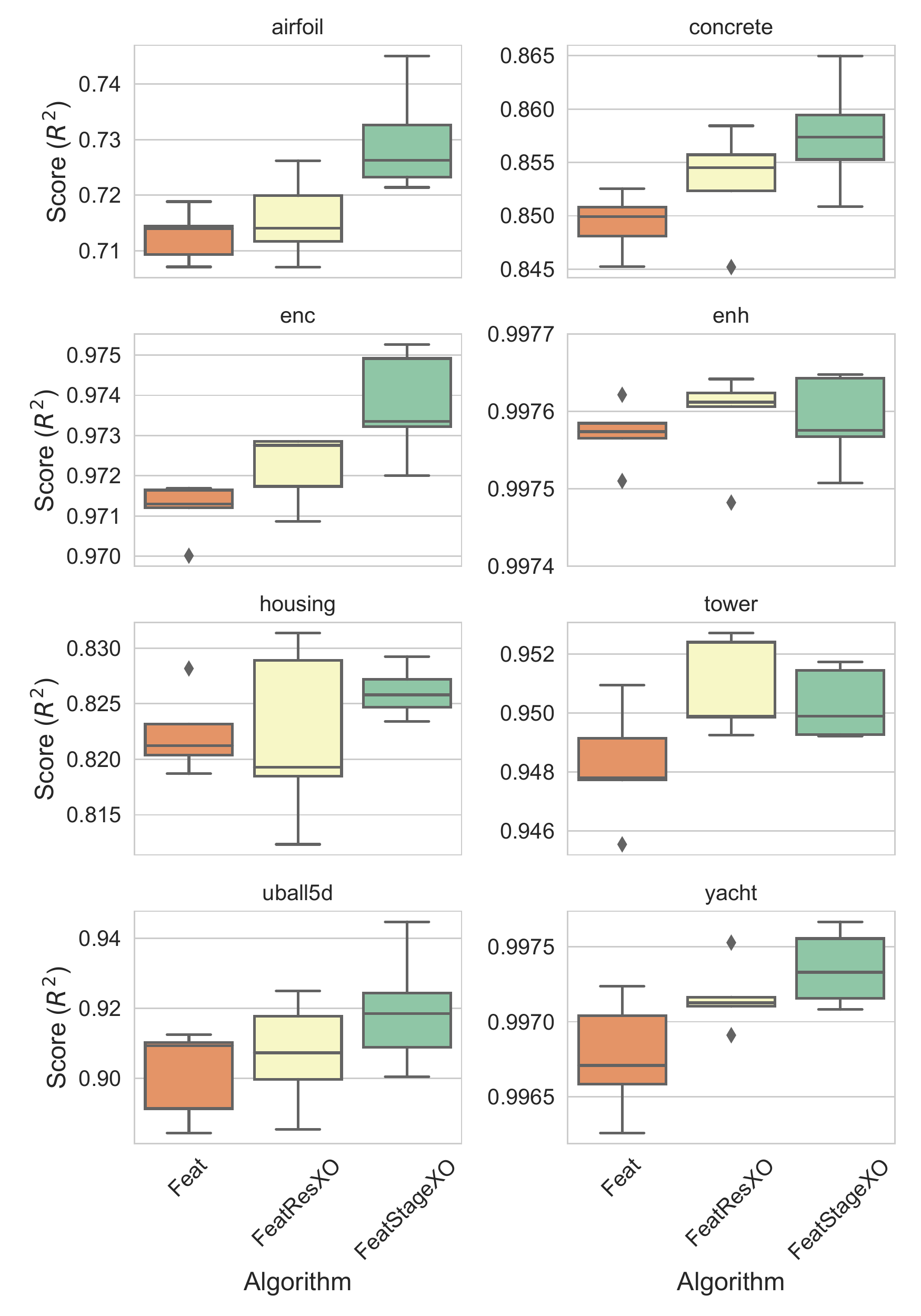}
    \caption{Mean 5-fold CV $R^2$ performance for different crossover operators on the 8 tuning problems.}\label{fig:score_tuning}
\end{figure}

Prediction comparisons for each crossover method are shown in Fig.~\ref{fig:score_tuning} for the 8 tuning problems. 
The plot shows the mean test fold $R^2$ value for the tuned estimator, summarized across trials.
In general one can see that StageXO produces the most accurate results, followed by ResXO. 
Across the 8 problems, StageXO significantly outperforms standard crossover ($p<$0.035); the pairwise statistical comparisons are given in Table~\ref{tbl:tuning_stats}.

\begin{figure}
    \centering
    \includegraphics[width=0.75\columnwidth]{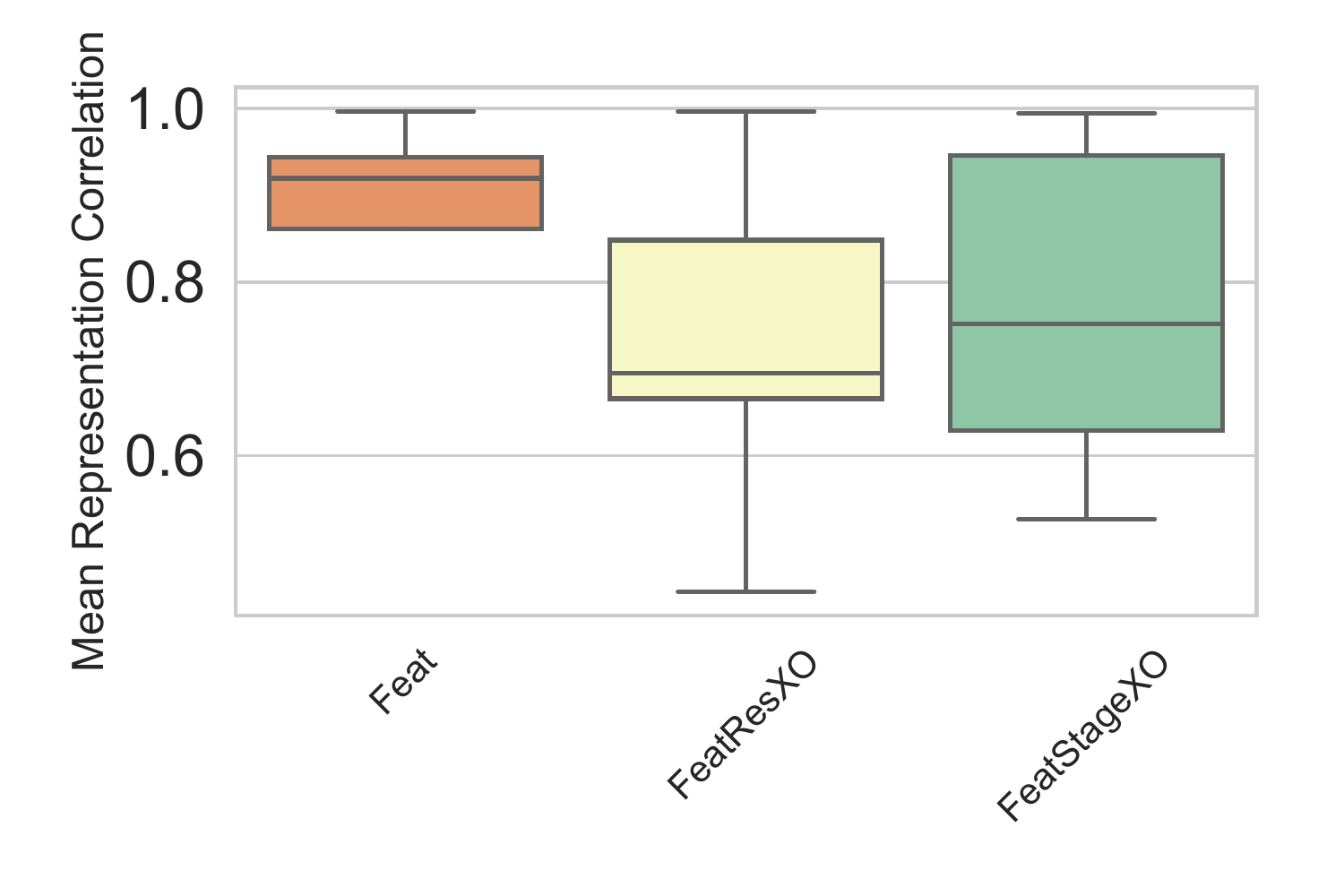}
    \caption{Average pairwise representation correlation (Eqn.~\ref{eq:corr}) for different crossover operators on the 8 tuning problems.}\label{fig:corr}
\end{figure}

We also looked at the correlation of the representations produced by the different crossover methods, shown in Fig.~\ref{fig:corr}. We confirmed our hypotheses that ResXO and StageXO would produce less correlated representations than the traditional crossover operator. 

\begin{figure}
    \centering
    \includegraphics[width=\columnwidth]{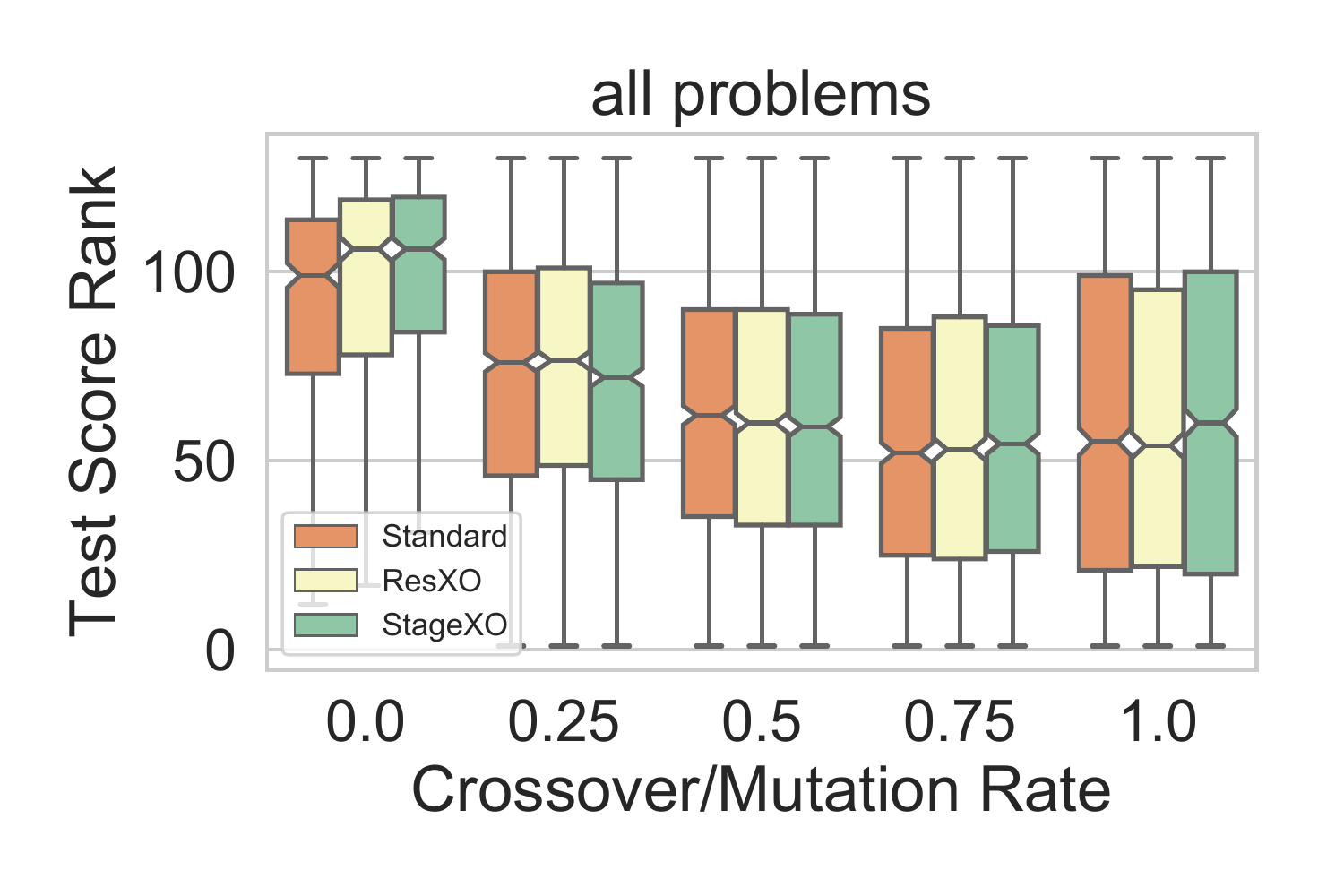}
    \caption{Test rankings for different crossover probabilities on the 8 tuning problems.}\label{fig:score_crossrate}
\end{figure}

\begin{figure}
    \centering
    \includegraphics[width=\columnwidth]{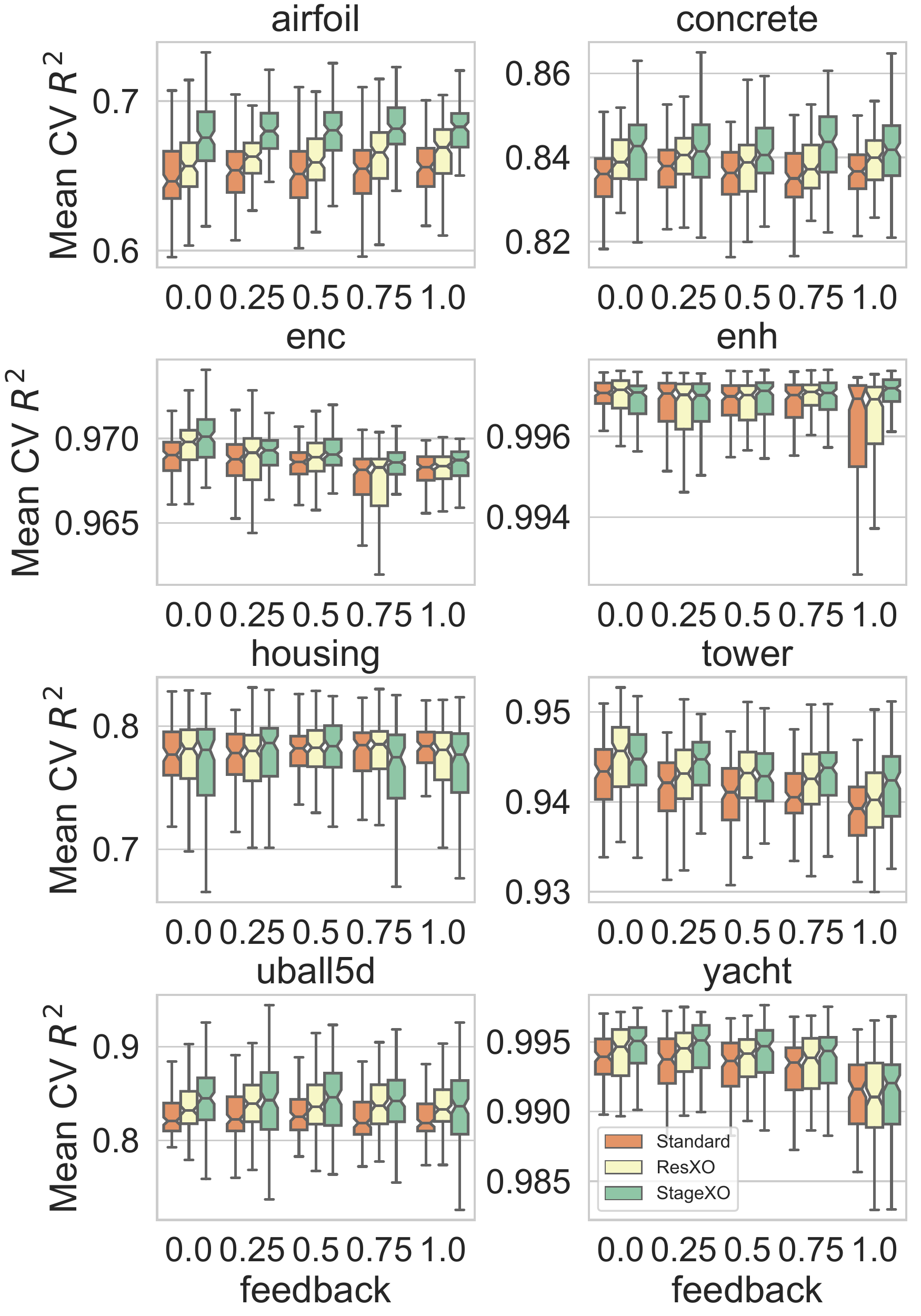}
    \caption{Mean 5-fold CV $R^2$ performance for different levels of feedback on the 8 tuning problems.}\label{fig:score_feedback}
\end{figure}

The best values for each tuned parameter is shown in Table~\ref{tbl:tuning}. 
We found that softmax normalization did not improve the feedback probabilities. 
Across problems, the best crossover/mutation fraction was found to be 0.75 (Fig.~\ref{fig:score_crossrate}), with a feature crossover rate of 0.75 for Feat and 0.5 for ResXO and StageXO. 
The best feedback value was problem dependent, as shown in Fig.~\ref{fig:score_feedback}. 
Since the feedback essentially controls the amount of exploration versus exploitation, it stands to reason that the ideal setting of this parameter would be problem dependent. 
Feedback levels of 0.25 were best for FEAT and FEATStageXO, and no feedback was best for FEATResXO. 
For the ResXO operator, this corresponds to choosing the feature to swap out of the parent at random.  

\begin{table}[ht]
\centering
\caption{Bonferroni-adjusted $p$-values using a Wilcoxon signed rank test of $R^2$ scores for the methods across all tuning problems. Bold: $p<$0.05.} \label{tbl:tuning_stats}
\begin{tabular}{rll}
  \hline
 & Feat & FeatResXO \\ 
  \hline
FeatResXO & 4.6e-01 &  \\ 
  FeatStageXO & {\bf 3.5e-02} & 1.2e-01 \\ 
   \hline
\end{tabular}
\end{table}

\begin{table}
    \footnotesize
    \centering
    \caption{Best hyperparameter values for FEAT across the 8 tuning problems.}\label{tbl:tuning}
    \begin{tabularx}{\columnwidth}{X r r r} \hline
        Hyperparameter & FEAT   & FEAT-ResXO    & FEAT-StageXO \\ \hline
        probability of crossover   &   0.75 & 0.75 & 0.75  \\
        feedback     &   0.25 & 0.0 & 0.25   \\
        probability of feature crossover  &   0.75 & 0.5 & 0.5    \\
        feedback softmax normalization  &   Off & Off & Off   \\
        \hline
    \end{tabularx}

\end{table}
\subsection{Benchmark comparison}
\begin{figure*}
    \centering
    \includegraphics[width=\textwidth]{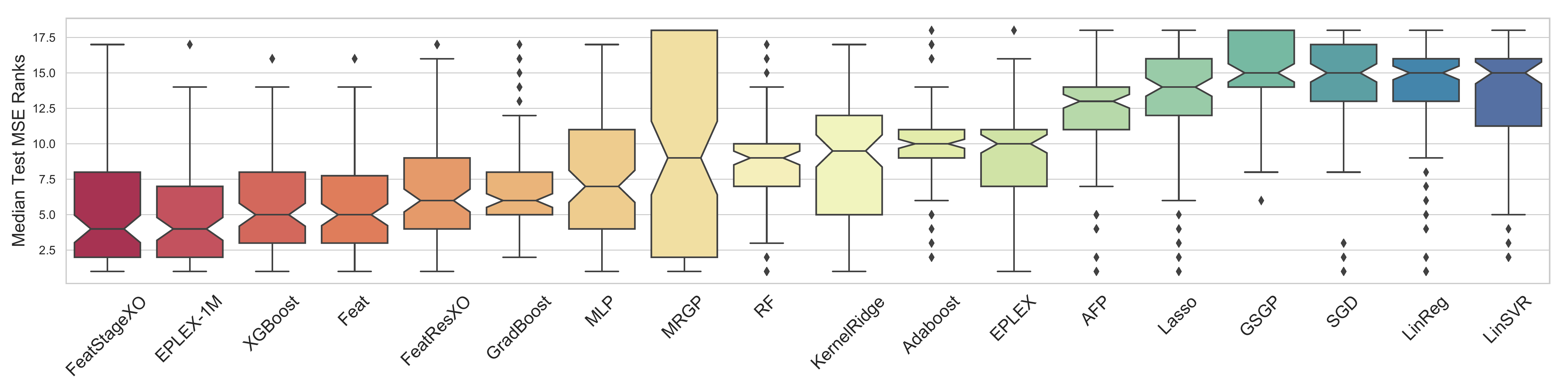}
    \caption{Median test MSE rankings on the PMLB datasets. The box shows the quartiles of the rankings with whiskers showing the rest of the distribution excluding outliers. Algorithms are ordered left to right by best (lowest) to worst (highest) median ranking.}\label{fig:score_bench}
\end{figure*}

\begin{figure*}
    \centering
    \includegraphics[width=\textwidth]{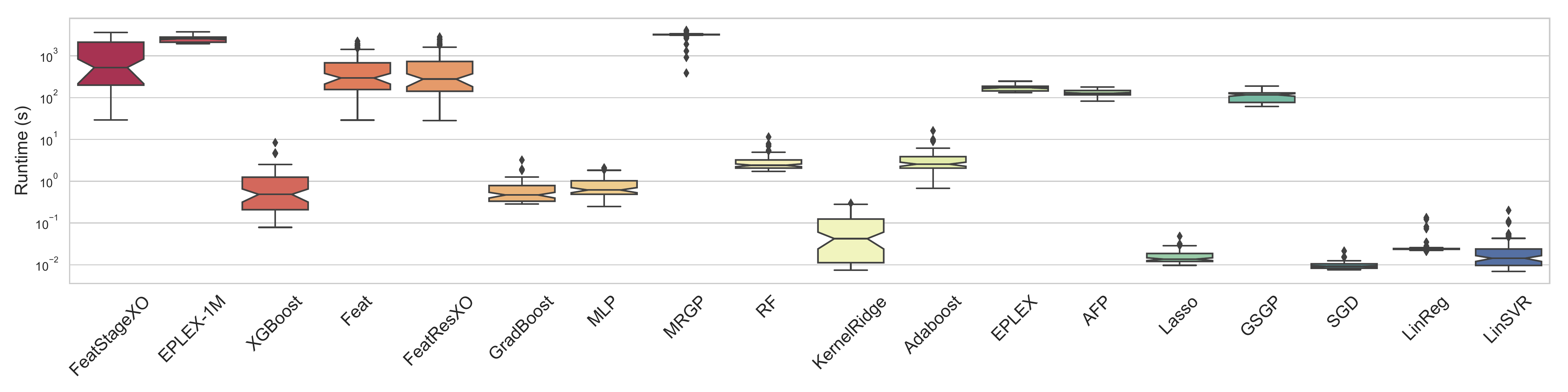}
    \caption{Runtime comparisons on the PMLB datasets. Runtime is the wall clock time for a single training instance. Algorithms are ordered to match Fig.~\ref{fig:score_bench}.}\label{fig:time}
\end{figure*}

The comparisons of FEAT to 15 other methods is shown in Fig.~\ref{fig:score_bench}. 
Across problems, FEATStageXO achieves a nearly identical ranking to EPLEX-1M, which is $\epsilon$-lexicase selection run for 1 million evaluations. In this case, FEATStageXO achieves these similar results using 100,000 evaluations. However, the additional complexity of fitting ML models to each individual makes the evaluation of each individual in FEAT more costly. The wall clock times shown in Fig.~\ref{fig:time} reflect this, as the FEAT  wall clock times sit somewhere between the methods that ran for 100,000 evaluations (GSGP, AFP, MRGP, EPLEX) and EPLEX-1M. 

A Friedman test of the MSE rankings across problems indicates significant differences. 
Table~\ref{tbl:stats} shows post-hoc pairwise Wilcoxon signed rank tests of the results. 
FEAT, FEATResXO, and FEATStageXO all significantly outperform the other GP-based methods run for 100,00 evaluations.
There is no significant difference found between the FEAT variants, EPLEX-1M, XGBoost, GradBoost, or MLP. 
The FEAT variants significantly outperform all other ML methods across the benchmark problems.

\begin{figure}
    \includegraphics[width=\columnwidth]{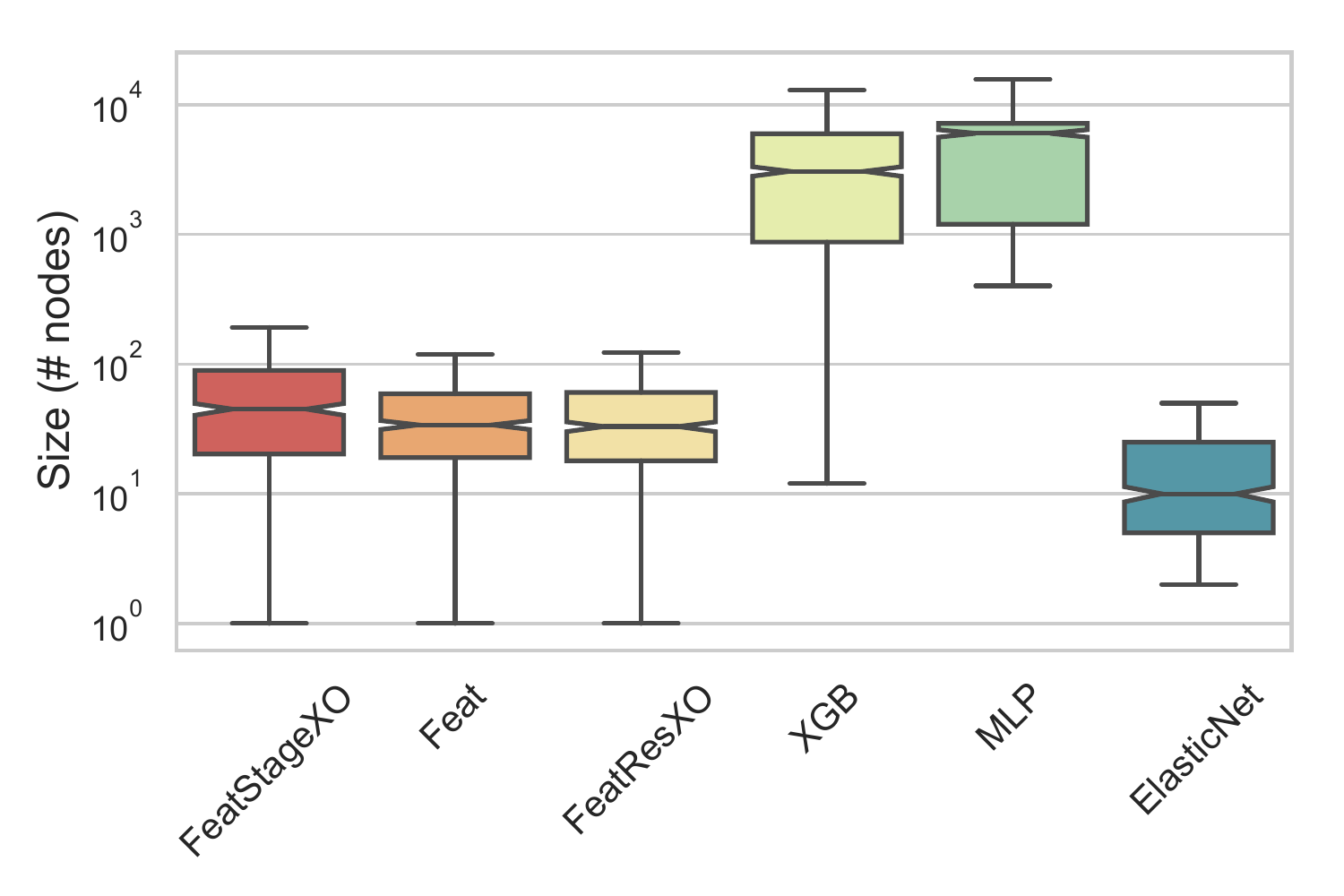}
    \caption{Number of nodes in solutions on the PMLB datasets.}\label{fig:size}
\end{figure}

We extend the comparison of FEAT to XGBoost, MLP and ElasticNet on 111 of the datasets in PMLB in order to evaluate the complexity of the final representations. 
The size comparisons are shown in Fig.~\ref{fig:size}. In general FEAT produces representations about 1.5 orders of magnitude smaller than XGBoost and MLP. 
We found that StageXO led to slightly larger models than Feat or FeatResXO.  

\begin{table*}[ht]
\tiny
\centering
\begin{tabularx}{\textwidth}{r X X X X X X X X X X X X X X X X X}
  \hline
 & Adaboost & AFP & EPLEX & EPLEX-1M & FEAT & FEAT-ResXO & FEAT-StageXO & GradBoost & GSGP & KernelRidge & Lasso & LinReg & LinSVR & MLP & MRGP & RF & SGD \\ 
  \hline
AFP & {\bf 1.2e-02} &  &  &  &  &  &  &  &  &  &  &  &  &  &  &  &  \\ 
  EPLEX & 1.0e+00 & {\bf 7.3e-08} &  &  &  &  &  &  &  &  &  &  &  &  &  &  &  \\ 
  EPLEX-1M & {\bf 8.2e-09} & {\bf 9.4e-12} & {\bf 3.4e-09} &  &  &  &  &  &  &  &  &  &  &  &  &  &  \\ 
  FEAT & {\bf 2.6e-08} & {\bf 5.8e-10} & {\bf 1.9e-06} & 1.0e+00 &  &  &  &  &  &  &  &  &  &  &  &  &  \\ 
  FEATResXO & {\bf 1.8e-04} & {\bf 2.2e-06} & {\bf 5.8e-04} & {\bf 5.0e-02} & {\bf 1.7e-02} &  &  &  &  &  &  &  &  &  &  &  &  \\ 
  FEATStageXO & {\bf 9.1e-08} & {\bf 8.2e-10} & {\bf 4.1e-06} & 1.0e+00 & 1.0e+00 & {\bf 5.8e-03} &  &  &  &  &  &  &  &  &  &  &  \\ 
  GradBoost & {\bf 1.5e-08} & {\bf 2.3e-08} & {\bf 5.6e-03} & {\bf 3.7e-02} & 1.0e+00 & 1.0e+00 & 1.0e+00 &  &  &  &  &  &  &  &  &  &  \\ 
  GSGP & {\bf 8.4e-11} & {\bf 2.6e-07} & {\bf 4.1e-12} & {\bf 1.5e-14} & {\bf 8.1e-14} & {\bf 5.1e-13} & {\bf 2.3e-14} & {\bf 2.6e-14} &  &  &  &  &  &  &  &  &  \\ 
  KernelRidge & 1.0e+00 & {\bf 1.5e-03} & 1.0e+00 & {\bf 8.1e-04} & {\bf 2.1e-02} & 1.0e+00 & {\bf 1.1e-02} & 9.9e-01 & {\bf 1.4e-14} &  &  &  &  &  &  &  &  \\ 
  Lasso & {\bf 8.3e-04} & 1.0e+00 & {\bf 1.9e-06} & {\bf 4.4e-12} & {\bf 4.5e-10} & {\bf 3.5e-07} & {\bf 4.8e-10} & {\bf 1.8e-08} & {\bf 2.1e-02} & {\bf 8.6e-08} &  &  &  &  &  &  &  \\ 
  LinReg & {\bf 1.1e-04} & {\bf 7.0e-03} & {\bf 2.3e-07} & {\bf 3.7e-12} & {\bf 4.5e-11} & {\bf 4.5e-09} & {\bf 6.1e-11} & {\bf 1.7e-08} & 1.0e+00 & {\bf 3.0e-09} & 1.0e+00 &  &  &  &  &  &  \\ 
  LinSVR & {\bf 1.1e-03} & 8.6e-02 & {\bf 2.3e-06} & {\bf 4.6e-12} & {\bf 1.3e-09} & {\bf 5.7e-07} & {\bf 9.0e-10} & {\bf 2.6e-08} & 3.9e-01 & {\bf 7.9e-09} & 1.0e+00 & 1.0e+00 &  &  &  &  &  \\ 
  MLP & {\bf 2.6e-02} & {\bf 2.6e-06} & 1.0e+00 & 1.0e+00 & 1.0e+00 & 1.0e+00 & 1.0e+00 & 1.0e+00 & {\bf 9.6e-15} & 7.4e-01 & {\bf 6.8e-07} & {\bf 2.0e-08} & {\bf 1.3e-07} &  &  &  &  \\ 
  MRGP & 1.0e+00 & 1.0e+00 & 1.0e+00 & {\bf 1.1e-04} & {\bf 7.0e-04} & 2.8e-01 & {\bf 3.6e-04} & 2.1e-01 & {\bf 2.1e-05} & 1.0e+00 & 2.3e-01 & {\bf 2.5e-02} & 1.5e-01 & 1.0e+00 &  &  &  \\ 
  RF & {\bf 7.9e-05} & {\bf 1.2e-04} & 1.0e+00 & {\bf 7.4e-06} & {\bf 1.0e-04} & 2.0e-01 & {\bf 7.6e-05} & {\bf 2.3e-06} & {\bf 2.1e-13} & 1.0e+00 & {\bf 4.7e-06} & {\bf 4.4e-06} & {\bf 1.3e-06} & 1.0e+00 & 1.0e+00 &  &  \\ 
  SGD & {\bf 2.0e-06} & {\bf 6.4e-05} & {\bf 1.9e-09} & {\bf 9.6e-13} & {\bf 2.7e-12} & {\bf 1.5e-09} & {\bf 1.3e-11} & {\bf 4.1e-11} & 1.0e+00 & {\bf 2.6e-09} & 5.5e-02 & 1.0e+00 & 1.0e+00 & {\bf 1.8e-09} & {\bf 4.5e-03} & {\bf 2.8e-09} &  \\ 
  XGBoost & {\bf 2.1e-08} & {\bf 6.9e-11} & {\bf 5.3e-05} & 1.0e+00 & 1.0e+00 & 1.0e+00 & 1.0e+00 & {\bf 8.9e-03} & {\bf 6.0e-15} & {\bf 6.9e-03} & {\bf 1.3e-10} & {\bf 6.4e-10} & {\bf 2.2e-10} & 1.0e+00 & {\bf 2.8e-03} & {\bf 3.6e-08} & {\bf 5.8e-12} \\ 
   \hline
\end{tabularx}
\caption{Bonferroni-adjusted $p$-values using a Wilcoxon signed rank test of MSE scores for the methods across all benchmarks. Bold: $p<$0.05.} \label{tbl:stats}
\end{table*}
\section{Discussion \& Conclusion}

This paper proposes semantic crossover operators for multidimensional genetic programming. 
We contend that MGP provides an interesting framework for developing semantic variation methods, due to the unique way that program semantics are easily teased apart in the multi-output architecture. 
The most successful of these is forward stagewise crossover (StageXO), which mimics forward stagewise selection to incrementally build an offspring from the representations of its parents. 
A lightweight version of semantic backpropagation crossover, ResXO, is also proposed. 
StageXO achieves better performance on 8 regression problems than naive crossover operators, taking into account hyperparameter tuning. 
We find that the resultant representations are also less ``entangled", i.e. less correlated when using these crossover methods.

We consider this to be a first foray into semantic methods for MGP, and consider the potential for more powerful techniques to be high. 
One can imagine many more semantic operators that take advantage of this architecture. 
Three extensions come to mind. 
The first is to consider the features of many programs during crossover, rather than just two parents, as is done with population wide semantic crossover~\cite{vanneschi_psxo:_2017}, behavioral GP, or in recent work by Fine et. al.~\cite{fine_exploiting_2018}. 
The second extension would be to explicitly design variation operators that improve the condition of the representation, for example by pruning highly correlated features. 
A third extension is to implement smarter termination conditions for stagewise crossover that take into account the fitness of the parents or an error tolerance. 

The current methods appear to be competitive with state-of-the-art regression techniques. 
In particular, FEAT is able to achieve top ranking results with less computational complexity than the previous top GP method on the PMLB benchmark. 
It is also able to produce much smaller representations than XGBoost and MLP. 
These promising results should motivate further research within the MGP framework.

\section{Supplementary Material}
Code to reproduce the experiments can be found at \url{http://github.com/lacava/gecco_2019}.

\section{Acknowledgments}
This work was supported by NIH grants AI116794 and LM012601, as well as the PA CURE grant from the Pennsylvania Department of Health. 
Special thanks to Tilak Raj Singh and members of the Computational Genetics Lab.

\bibliographystyle{ACM-Reference-Format}
\bibliography{gecco_feat} 

\end{document}